\documentclass[lettersize,journal]{IEEEtran} 

\IEEEoverridecommandlockouts                              

\usepackage{graphicx}
\usepackage{amsmath,amssymb}
\usepackage{cite}
\usepackage{float}
\usepackage[caption=false,font=footnotesize]{subfig}
\usepackage{algorithm}
\usepackage{algorithmic}
\usepackage{color}
\definecolor{review_color}{rgb}{0,0,0}

\graphicspath{{img/},{fig/}}

\renewcommand{\vec}[1]{\boldsymbol{#1}}

\begin{document}

\title{
Leveraging Natural Load Dynamics with Variable Gear-ratio Actuators
}


\author{Alexandre Girard$^1$ and H. Harry Asada$^2$
\thanks{This work was supported by The Boeing Company. }%
\thanks{$^1$ A. Girard is with the Department of Mechanical Engineering, Universite de Sherbrooke, Qc, Canada.} 
\thanks{$^2$ H. H. Asada is with Department of Mechanical Engineering, Massachusetts Institute of Technology, Cambridge, MA, USA.}%
\thanks{$^3$ \textcopyright IEEE. Personal use of this material is permitted. Permission from IEEE must be obtained for all other uses, in any current or future media, including reprinting/republishing this material for advertising or promotional purposes, creating new collective works, for resale or redistribution to servers or lists, or reuse of any copyrighted component of this work in other works. DOI:10.1109/LRA.2017.2651946}
}

\markboth{IEEE Robotics and Automation Letters,~Vol.~2, No.~2, April~2017, Preprint version. DOI:10.1109/LRA.2017.2651946$^3$}{}

\maketitle

\begin{abstract}
This paper presents a robotic system where the gear-ratio of an actuator is dynamically changed to either leverage or attenuate the natural load dynamics. Based on this principle, lightweight robotic systems can be made fast and strong; exploiting the natural load dynamics for moving at higher speeds (small reduction ratio), while also able to bear a large load through the attenuation of the load dynamics (large reduction ratio). A model-based control algorithm to automatically select the optimal gear-ratios that minimize the total actuator torques for an arbitrary dynamic state {\color{review_color} and expected uncertainty level is proposed}. Also, a novel 3-DoF robot arm using custom actuators with two discrete gear-ratios is presented. The advantages of gear-shifting dynamically are demonstrated through experiments and simulations. Results show that actively changing the gear-ratio using the proposed control algorithms can lead to an order-of-magnitude reduction of necessary actuator torque and power, and also increase robustness to disturbances. 
\end{abstract}



\section{INTRODUCTION}

The transmission gear-ratio that couples an actuator to a load has a significant effect upon the behavior of the actuator-load system. With a large reduction ratio, the load-side dynamics has no significant effect because it is attenuated by the factor of the square of the gear-ratio. The net load acting on the actuator is mostly its own intrinsic load, including rotor inertia and friction. In contrast, with a small reduction ratio or a direct drive system \cite{asada_direct-drive_1987}, the behavior is usually dominated by the load-side dynamics which consist of highly non-linear inertial and gravitational forces for robotics manipulators. Sometime it can be advantageous to exploit the load-side dynamics: gravity may push the robot in a desired direction; the robot may coast with small dissipative torques induced at the actuator side; or the robot joints become backdrivable to comply to an external force. In other situations, however, it may be advantageous to isolate the actuators from the load-side dynamics and external disturbances: using a large gear-ratio to bear a large load or moving it slowly against gravity, for example.

This paper aims to explore the potentials of actuator transmissions that can be switched dynamically to either attenuate or leverage the natural dynamics of the system. {\color{review_color} Here, the variable gear-ratio is used not merely for increasing maximum torque and speed, but also to significantly alter the dynamic properties, including load sensitivity, robustness, and backdrivability. 
Exploiting the full potentials of variable gear-ratios, with effective gearshift algorithms, entails a sound understanding of the effect of actuator gear-ratios in diverse static and dynamic robotic contexts.} In section \ref{sec:princ}, the principle of load leveraging and attenuation is delineated for a simple 1-DoF manipulator, section \ref{sec:app} will discuss how this principle can be exploited in various contexts of robotic applications, where bi-polar load conditions must be dealt with despite actuator weight and power constraints, and section \ref{sec:chal} will discuss related works and technical challenges. Section \ref{sec:ctl} will introduce a formal mathematical representation and propose control algorithms to automatically select optimal gear-ratios for a single DoF and multi-DoF systems. The advantage of actively changing the gear-ratio are then illustrated with simulations in section \ref{sec:sim}, and with experiments using a custom robotic arm in section \ref{sec:ev}.

\subsection{Illustration of the principle for a 1-DoF manipulator}
\label{sec:princ}

\begin{figure}[H]
	\centering
		\includegraphics[width=0.47\textwidth]{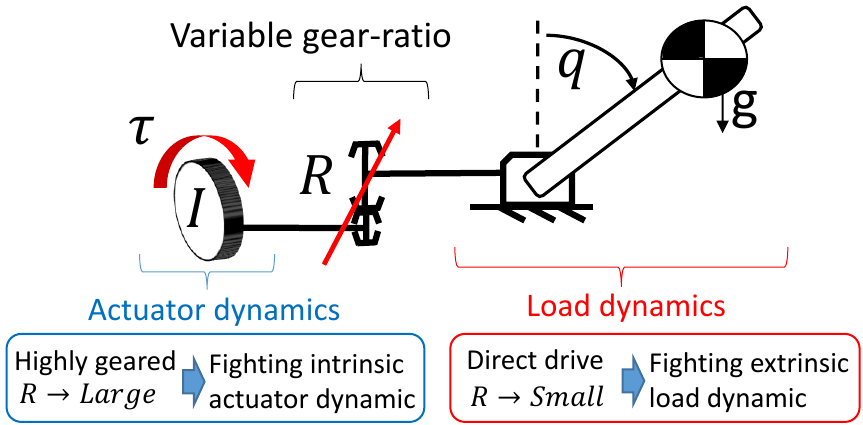}
	\caption{Effect of the gear-ratio on the natural dynamics}
	\label{fig:bigpicture}
\end{figure}

Fig. \ref{fig:bigpicture} illustrates a simplified 1-DoF robotic manipulator where an electric motor is coupled to a pendulum through a gearbox with a variable gear-ratio $R$.

\begin{figure}[H]
				\vspace{-10pt}
        \centering
				\subfloat[ Gear-ratio $R$=1 ]{ 
				\includegraphics[width=0.24\textwidth]{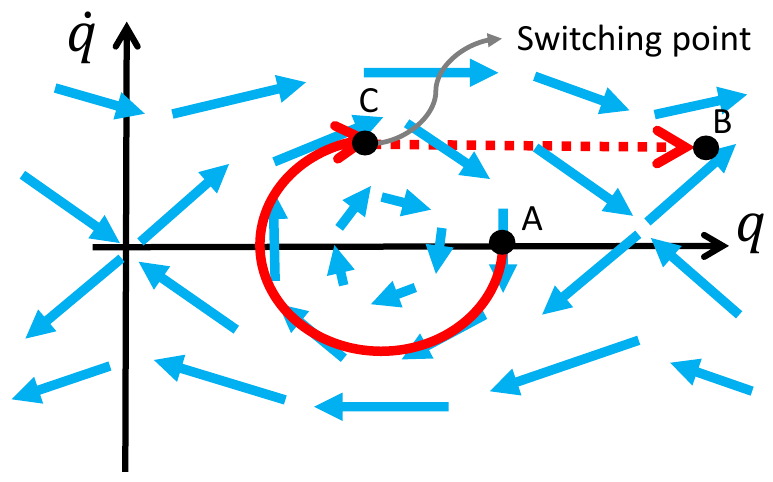}
				\label{fig:pp1}}
        \subfloat[ Gear-ratio $R$=10 ]{ 
				\includegraphics[width=0.23\textwidth]{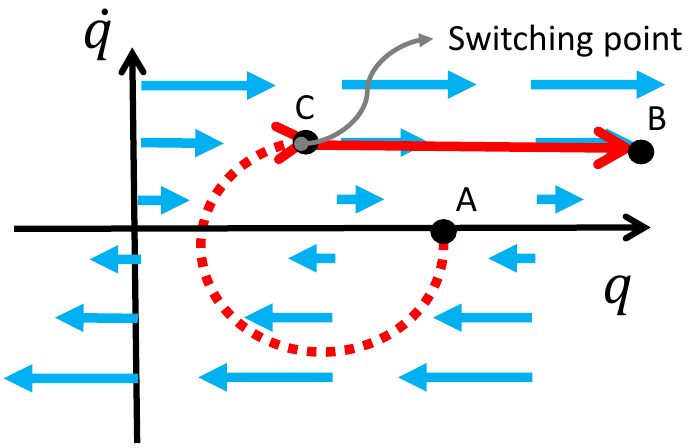}
				\label{fig:pp2}}
        \caption{Phase portraits illustrating the dynamical behavior}
				\label{fig:pp}
\end{figure}

As illustrated by phase portraits in Fig. \ref{fig:pp}, if $R$ is small then the dynamic behavior of the system is dominated by the non-linear pendulum dynamics (Fig. \ref{fig:pp1}), but if $R$ is very large, the behavior is dominated by the intrinsic inertia of the actuator, leading to the double-integrator behavior (Fig. \ref{fig:pp2}).

The vector fields of Fig. \ref{fig:pp} illustrate the evolution of the system with no actuator torque. Suppose that we want to move from state A to state B on the phase plane. Starting off state A with the gear-ratio of 1:1 brings the system along the curved trajectory shown in Fig. \ref{fig:pp1}. Switching the gear-ratio at state C to 1:10 will change the trajectory to the one in Fig. \ref{fig:pp2}, and bring the system to the destination state B. Note that no actuator torque is necessary for following this trajectory. A salient feature of actively changing the gear-ratio is that a natural behavior, i.e. vector field, having properties useful for moving in a desired direction can be selected to minimize the necessary torque to apply to the system.

\subsection{Diverse Load Conditions}
\label{sec:app}

\begin{figure}[h]
	\centering
		\includegraphics[width=0.48\textwidth]{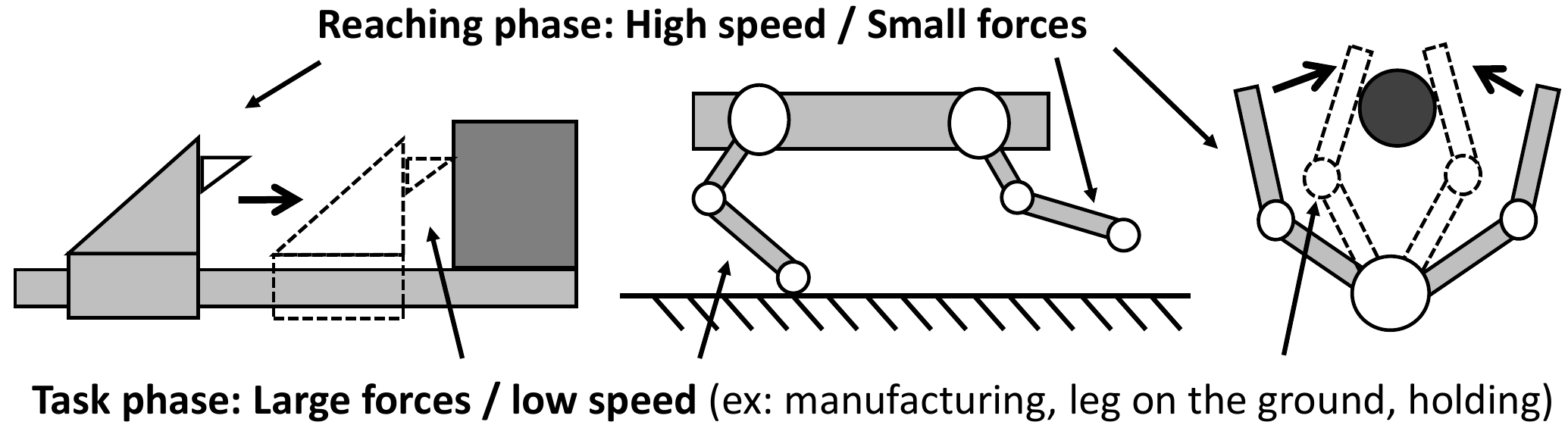}
	\caption{Robotic systems encountering diverse load conditions; the load is significantly small when the end-effector, legs, and fingers are moving quickly through the air, while they have to bear large loads when in contact with the environment or moving against gravity.}
	\label{fig:app2}
\end{figure}

In many applications, as illustrated in Fig. \ref{fig:app2}, robotic systems encounter a very wide range of load conditions: fast reaching motions where forces are small (small gear-ratio is advantageous), and applying large forces at low speed (large gear-ratio is advantageous).


\begin{figure}[h]
	\centering
		\includegraphics[width=0.45\textwidth]{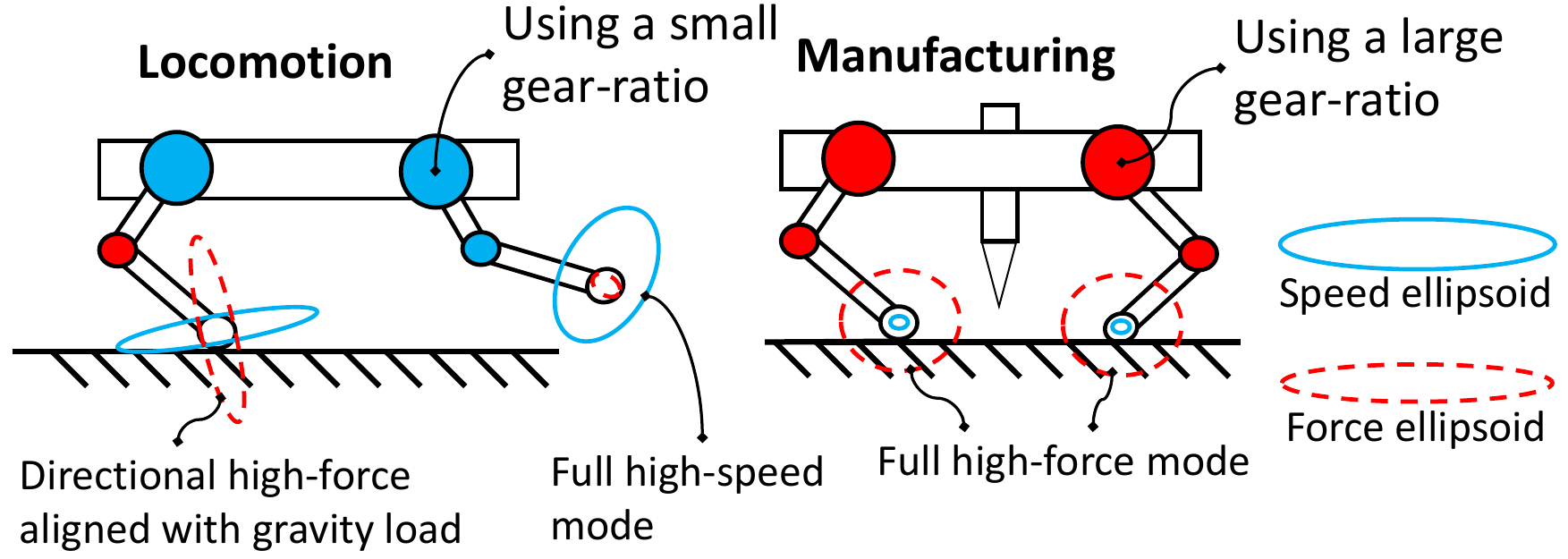}
	\caption{Example of advantageous gear selections with a multi-DoF robot}
	\label{fig:gearselectionlegged}
\end{figure}

{\color{review_color}Lightweight small motors can meet those requirements when equipped with a variable gear reducer \cite{girard_two-speed_2015}.} In addition to weight-saving, advantages of variable gear-ratio actuators (VGA) include improved efficiency and the wide range of possible impedance reflected to the load side \cite{vanderborght_variable_2013}. As illustrated in Fig. \ref{fig:gearselectionlegged}, a leg can be made fast and backdrivable by selecting small gear-ratios, or slow and capable of bearing a large load with a large gear-ratio. Furthermore, a leg can be made stiff in one direction but fast in other directions by selecting different gear-ratios for different axes.

 {\color{review_color} Robots using lightweight VGA have the potential for achieving fast motions, high load-bearing, compliance and high-impedance, as required by diverse load conditions encountered by robotic systems. However, to truly exploit those salient features of VGA, control laws to select dynamically gear-ratios based on the current situation and task of the robot must be developed. }



{ \color{review_color}
}


\subsection{Challenges and related works}
\label{sec:chal}


This paper investigates the control schemes for robots using VGA, focusing on intelligent gear-ratio selection. VGA differs from variable stiffness actuators (VSA) that use a variable transmission placed between a compliant element and the load \cite{vanderborght_variable_2013}. {\color{review_color} Both VSA and VGA can vary the impedance reflected to the load side, but only VGA has the advantage of improving power transmission and efficiency by matching motor speed to the load speed (like a car transmission). On the other hand, only VSA have the ability to store and release potential energy in their compliant element.} While VGA have been studied extensively for automobile power-trains, they have not yet been fully investigated in robotics, despite significant potential gains. Variable gear-ratio transmissions for electric motors have been proposed for legged locomotion \cite{hirose_design_1991}, grasping robotic hands \cite{shin_robot_2012} , propulsion system \cite{lee_new_2012} \cite{mckeegan_antonovs_2011} and actuation systems \cite{girard_two-speed_2015} \cite{hirose_development_1999} \cite{tahara_high-backdrivable_2011}. Some of those works address the issue of how to change the gear-ratio, but there is no general approach to the high-level control of automatically selecting the right gear-ratios for nonlinear, coupled multi-DoF robotic systems.

From the control perspective, automating the gear-ratio selection in a robotic context is a new and challenging problem. Gear-shifting is a very non-linear process and the plant becomes a hybrid dynamical system if the usable gear-ratios are a set of discrete values. In simple scenarios, the gear-ratio selection can be based on simple principles. For instance, for a system running at a steady speed, the best gear-ratio can be selected based on efficiency. Alternatively, for rapid acceleration, the gear-ratio may be selected based on the actuator-load inertia matching \cite{giberti_effects_2010} \cite{chen_generalized_1991}. Robots, however, experiences diverse types of forces acting simultaneously. These include gravity, friction, and inertial forces as well as Coriolis and centrifugal forces. Hence, it is challenging to find a general control policy for selecting gear-ratios for the multitude of dynamically interacting actuators in the context of robotic systems. {\color{review_color} 

Mixed-integer programming has been used to generate optimal open-loop trajectories of a dynamical system with both a continuous torque and a discrete gear-selection input variable \cite{gerdts_solving_2005}. However, open-loop trajectories can be unstable and computation time is prohibitive. Dynamic programming was used to generate feedback laws for torque and gear-ratio selection for a 1-axis system \cite{girard_practical_2016}. However, this technique does not scale to high-dimensional robots and is not applicable to trajectory following. Here in this paper, a model-based approach is proposed, with the advantage of: scaling to high-dimensional robotic systems, adapted for trajectory following tasks and can use a measure of uncertainty for improved robustness. Given a robot model, a reference trajectory, and optionally known bounds on the uncertainty, feedback laws for gear-ratios selection and motor torques are synthesized.} The focus will be on systems where the gear-ratio options are limited to a discrete set, but the proposed principle and control algorithms are also applicable to robots with continuously variable transmissions (CVT).




\section{Control }
\label{sec:ctl}

\begin{figure}[htp]
	\centering
		\includegraphics[width=0.31\textwidth]{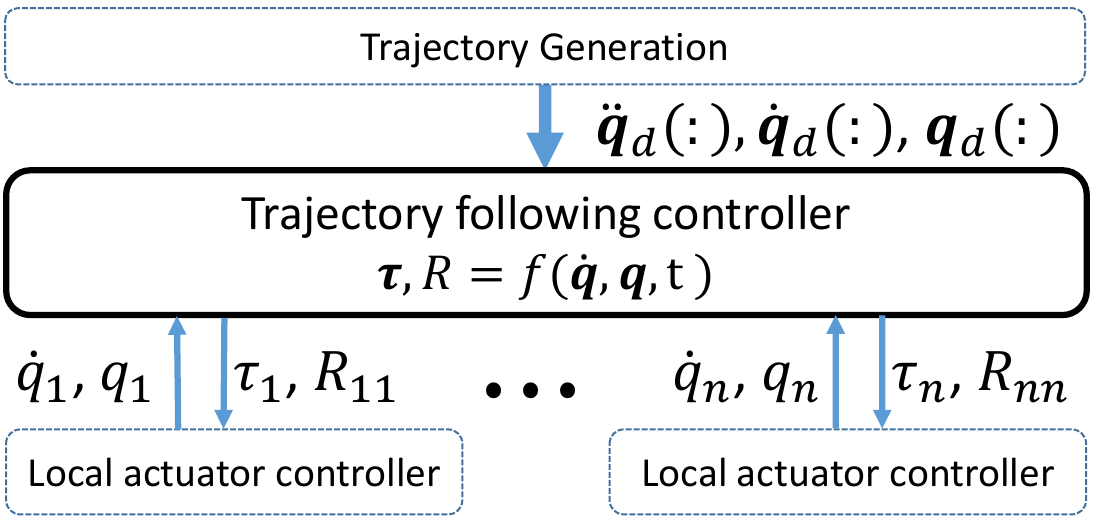}
	\caption{Proposed hierarchical control architecture}
	\label{fig:control_achitecture}
\end{figure}

The proposed control architecture, shown in Fig. \ref{fig:control_achitecture}, consists of three hierarchical control loops. First, a trajectory generation algorithm synthesizing dynamic trajectories that meets performance requirements for reaching desired states. Second, a closed-loop trajectory following controller consisting of a feedback law computing actuator torques $\vec{\tau}$ and gear-ratios $R$ based on the measured full state of the robot. Finally at the lowest level, independent actuator controllers executing low-level hardware commands in response to given torques and gear-ratio set-points. 

The main scope of this paper is the trajectory following controller: the design of a feedback law optimizing gear-ratios and computing torque commands in real-time. For the trajectory generation, motion planning algorithms can be applied to optimize the overall trajectory \cite{lavalle_planning_2006}.  The low-level controllers would be specific to the type of actuator used in the system. It is assumed that they would track the desired torque/force and handle the gear-shifting process when receiving a new gear-ratio set-point. 


\subsection{Modeling}
\label{sec:mod}

%
\begin{table}[tbp]
	\centering
	\vspace{+10pt}
		\begin{tabular}{ c c l }
        \hline \hline
			$H$             &  :  & External inertia matrix \\
			$D$             &  :  & External damping matrix \\
			$C$             &  :  & External Coriolis/Centrifugal forces matrix  \\
			$\vec{g}$       &  :  & External gravitational vector  \\
			$\vec{d  }$     &  :  & External disturbances vector  \\
			$R$             &  :  & Gear-ratio matrix  \\ 
			$I$             &  :  & Intrinsic actuator inertia matrix (diagonal) \\
			$B$             &  :  & Intrinsic actuator damping matrix (diagonal) \\
			$\vec{\tau}$    &  :  & Electromagnetic motor torques vector \\
			$\vec{q}$       &  :  & Joint coordinates position vector  \\
			$\vec{w}$       &  :  & Actuator coordinates velocity vector  \\
			$J$             &  :  & task-space coordinates / joint coordinates jacobian matrix\\
			$\vec{f}$       &  :  & Net transmitted forces vector (joint coordinates) \\
			$\vec{\tau}'$   &  :  & Net transmitted forces vector (actuators coordinates) \\
			$\vec{\tau}_I$  &  :  & Sum of intrinsic forces vector  \\
			$\vec{\tau}_E$  &  :  & Sum of extrinsic forces vector \\
		\hline \hline
        \end{tabular}		
        \caption{Nomenclature}	
		\vspace{-20pt}
	\label{tab:nom}
\end{table}
%
{\color{review_color} 
First, a generic 1-DoF robot with a variable transmission is considered for simplicity. If the actuator's intrinsic resistive forces $\tau_I$ are approximated to a linear quantity, the equations of motion (EoM) can be written as:}
\begin{align}
	\underbrace{\left[	H \ddot{q} + D \dot{q} + g( q )	\right]}_{\tau_{E}(\ddot{q},\dot{q},q)}
	&= R \tau - R^2
	\underbrace{\left[ I \ddot{q} + B \dot{q}	\right]}_{\tau_{I}(\ddot{q},\dot{q})} \\
	\tau &= 	\frac{\tau_{E}(\ddot{q},\dot{q},q)}{R} + R \; \tau_{I}(\ddot{q},\dot{q})
	\label{eq:1dofEoM}
\end{align}
where the effect of the gear-ratio can be seen clearly; increasing $R$ attenuates the external dynamic terms $\tau_{E}$ but amplify the intrinsic actuator losses $\tau_{I}$ for a given trajectory.

\begin{figure}[htp]
	\centering
		\includegraphics[width=0.27\textwidth]{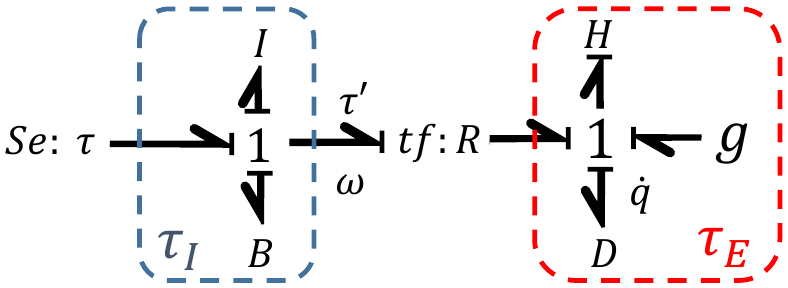}
	\caption{Model of a 1-DoF robot with a variable gear-ratio actuator}
	\label{fig:bondgraph1}
\end{figure}

Variable gear-ratios can be modeled as variable transformer elements, using the bond-graph terminology. Fig. \ref{fig:bondgraph1} shows a bond-graph representation of the model.

\begin{figure}[htp]
	\centering
		\includegraphics[width=0.43\textwidth]{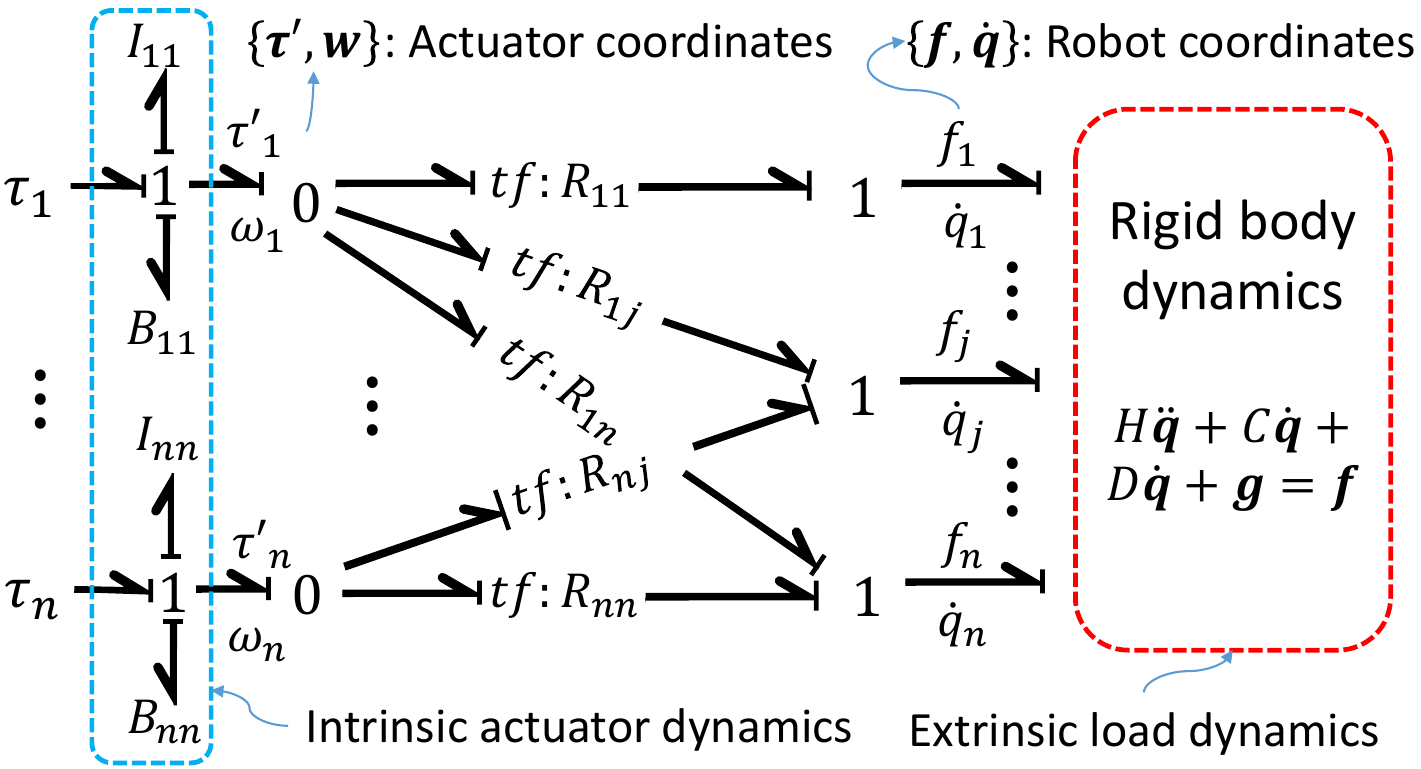}
	\caption{Model of a $n$-Dof robot with variable actuator-joint coupling}
	\label{fig:bondgraph}
\end{figure}

To generalize the above model to an $n$-DoF system with $n$ actuators, the load-side dynamics is considered as a generic form of manipulator dynamics where each port is connected to an independent motor through a network of transformers, as shown by Fig. \ref{fig:bondgraph}. The network of transformers can be view as a type of coordinate transformation relating effort (force or torque) and flow (velocity or angular velocity) on the load side ($\vec{f}$,$\dot{\vec{q}}$) to those on the actuator output side ($\vec{\tau}'$,$\vec{w}$):
\begin{align}
	\vec{ f } = R^T \vec{\tau}' \quad  \quad R \dot{ \vec{q} } = \vec{w}
 \label{eq:coortransform}
\end{align}
where $R$ is a $n$ by $n$ matrix consisting of all the transformer ratios. The EoM are then given by:
\begin{align}
	&\underbrace{ H \vec{ \ddot{q} } + C\vec{ \dot{q} } + D \vec{ \dot{q} } + \vec{g} }_{ \vec{\tau}_{E}(\ddot{\vec{q}},\dot{\vec{q}},\vec{q})}
		= R^T \underbrace{  \left[ 
		\vec{ \tau } - I \vec{ \dot{w} } - B \vec{ w }       
		\right]}_{ \vec{\tau}' } 
 \label{eq:eom_ndof}
\end{align}
Note that, in the case of locomotion or manipulation where the robot interacts with the environment by physically contacting it, the dynamic model $\vec{\tau}_{E}$ must reflect the contact conditions, either by computing contact forces using constraint equations or by formulating $\vec{\tau}_{E}$ as a hybrid dynamical system. 
In most practical cases, each actuator has its independent variable transmission and, thereby, the $R$ matrix will be diagonal and each diagonal value can be selected independently. Assuming this situation, the EoM can be simplified to a form, similar to the scalar case, illustrating the effect of the gear-ratios matrix $R$ on the behavior: 
\begin{align}
	\vec{\tau} &= R^{-1} 
	\underbrace{ 
	\vec{\tau}_{E}(\ddot{\vec{q}},\dot{\vec{q}},\vec{q} ) 
	}_{\text{External load dynamics}}
	+ R 
	\underbrace{ 
	\vec{\tau}_{I}(\ddot{\vec{q}},\dot{\vec{q}})
		}_{\text{Intrinsic losses}}
		\label{eq:eom_ndof2}
	\\ 
	\vec{\tau}_{I} &\triangleq I \vec{ \ddot{q} } + B \vec{ \dot{q} } 
\end{align}
%
%
%
The main assumption of the proposed model is that gear-ratios are considered as independent control inputs, neglecting all the dynamics and delays associated with the transition of gear-ratios. Physically this implies that the kinetic energy of the system may be discontinuous at a gear-shift since the energy necessary for the transition is not considered. In the case of a car transmission for instance, this model would not keep track of the energy used for accelerating or braking the engine during the synchronization process. This model can be used if the gear-shift process is fast compared to the dynamics of the robots and if the energetic losses due to the gear-shift are negligibly small. In addition this model also assumes that all motor rotors are in an inertial reference frame, neglecting gyroscopic effects, which may be induced when the axes of motor rotors are rotated.


\subsection{Optimal gear-ratio along a trajectory}

This section analyzes the optimal gear-ratio at each instant along a known trajectory. The two main advantages of changing gear-ratio are 1) lowering the necessary torque to follow a trajectory and 2) modifying the effective impedance reflected on the environment. Optimization for reducing torque can be done by minimizing $\vec{\tau}^T \vec{\tau}$ at each point along the trajectory. Optimization for reflected impedance can be done by minimizing the difference between desired task-space impedance and the actual one, which is directly affected by the matrix $R$. For instance, the end-point inertia matrix contains the gear-ratios: 
\begin{align}
	M = [J(\vec{q})^T]^{-1} \big [ \underbrace{ R^T I R }_{\text{Actuator contribution}} + H( \vec{q} ) \big ] J(\vec{q})^{-1}
 \label{eq:endpointmass}
\end{align}
Another point of practical importance is that rotor speeds should be constrained to be lower than their maximum velocity. This is to avoid infeasible gear shifts. For example, attempting to shift to a low gear at an extremely high speed is impossible. The best gear-ratios R*, according to those criteria, can be computed by solving the following optimization problem:
\begin{align}
	R^{*}(\ddot{\vec{q}},\dot{\vec{q}},\vec{q}) &= \operatornamewithlimits{argmin}\limits_{R} \left[ \vec{\tau}^T \vec{\tau} + \alpha \| M_{d} - M \| \right]  \\
	& \text{s.t.}  \quad R \dot{\vec{q}} \leq \vec{w}_{max} 
\label{eq:rmin_general}
\end{align}
where $\alpha$ is a parameter to set the trade-off between minimizing motor torques and matching the desired end-point inertia $M_d$.
Analytical solutions can be found when considering the minimal torque criterion only and linear intrinsic resistive forces. The optimal gear-ratio for a 1-DoF system, at a given instant on a trajectory, is given by
\begin{align}
	R^{*} &= \operatornamewithlimits{argmin}\limits_{R} \left[ \tau^2 \right] = \sqrt{ \left | \frac{\tau_{E}(\ddot{q},\dot{q},q)}{\tau_{I}(\ddot{q},\dot{q})} \right |   } 
\label{eq:aaa}
\end{align}
Similarly for multi-DoF systems, if $R$ is a diagonal matrix, the optimal gear-ratios can be obtained independently for each axis:
\begin{align}
	[R^*]_{ii} = \sqrt{ \left | \frac{ [\vec{\tau}_{E}(\ddot{\vec{q}},\dot{\vec{q}},\vec{q})]_i }{ [\vec{\tau}_{I}(\ddot{\vec{q}},\dot{\vec{q}})]_i } \right | }
 \label{eq:rmin2}
\end{align}

Note that large gravitational forces or contact forces, only present in $\vec{\tau}_{E}$, will usually lead to larger optimal gear-ratios, unless they cancel out other forces in a way that makes $\vec{\tau}_{E}$ smaller. If inertial or viscous forces, present both in $\vec{\tau}_{E}$ and $\vec{\tau}_{I}$, dominate, then the optimal gear-ratio will be a compromise such that extrinsic and intrinsic forces are balanced, a form of impedance matching. The optimal gear-ratio given by \eqref{eq:rmin2} includes both gravity, inertial and viscous forces as well as all other effects, hence it can be applied to any arbitrary dynamic situations.

\subsection{ \color{review_color} Uncertainty}
\label{sec:uncertainty}

{ \color{review_color}
Two observations are made regarding the effect of the gear-ratios on disturbances. First, considering modeling errors and external forces on the extrinsic side as unknown generalized forces $\vec{d}$, the EoM given by eq. \eqref{eq:eom_ndof2} becomes:
\begin{align}
	\vec{\tau} &= R^{-1} 
	\vec{\tau}_{E}(\ddot{\vec{q}},\dot{\vec{q}},\vec{q}) 
	+ R 
	\vec{\tau}_{I}(\ddot{\vec{q}},\dot{\vec{q}})
    + R^{-1}
    \underbrace{ 
	\vec{d}
	}_{\text{Disturbances}}    
 \label{eq:eom_ndof3}
\end{align}
where it is assumed that the actuators are accurately modeled. Note that the effect of the disturbances is inversely proportional to the gear-ratios, and thereby attenuated when using large gear-ratios. 

Second, large gear-ratios also decrease the sensitivity of the system to uncertainty. The error of computed accelerations will be attenuated with large gear-ratios because of the larger  actuator inertia reflected to the extrinsic side:
\begin{align}
	\vec{\ddot{q}}_e &= \vec{\ddot{q}} - \vec{\ddot{q}}_r = 
	\left[ 
    H + R^T I_a R
	\right]^{-1}
    \vec{d}
 \label{eq:sens}
\end{align}
Hence, selecting large gear-ratios makes the system less sensitive to uncertainty on the extrinsic side.
}

\subsection{Trajectory following control algorithm}
\label{sec:ctlalgo}

\begin{figure*}[t]
	\centering
		\includegraphics[width=0.95\textwidth]{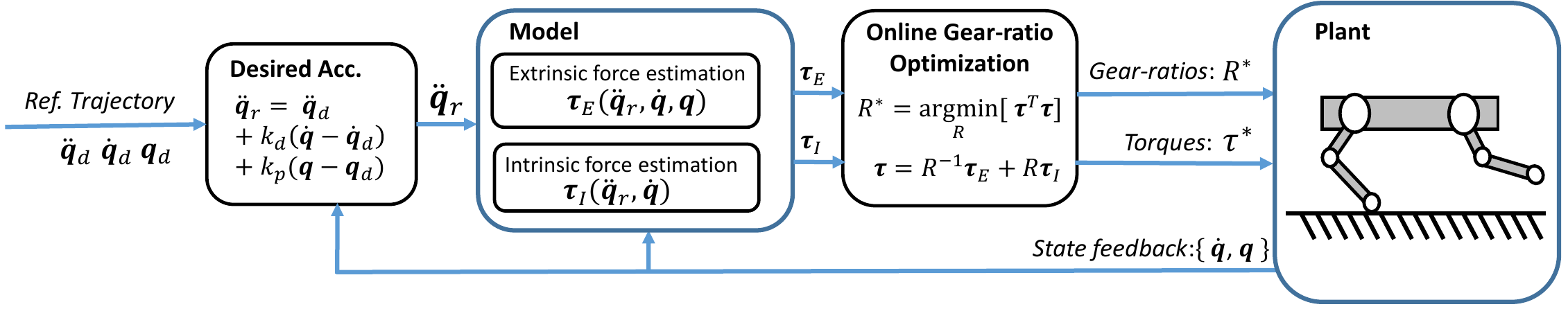}
	\caption{R* Computed Torque Controller illustrated with a block diagram}
	\vspace{-10pts}
	\label{fig:bbb}
\end{figure*}

This section proposes feedback laws, to actively select gear-ratios based on states and a reference trajectory.

\subsubsection{R* Computed Torque Control}
\label{sec:RComputedTorqueControl}

The proposed closed-loop controller, shown in Fig. \ref{fig:bbb} with a block diagrams, is based on the Computed Torque Control technique \cite{asada_robot_1986}, but includes an optimization step to compute and select the optimal gear-ratios. The idea is as follow, first compute a necessary acceleration $\ddot{\vec{q}}_r$ to guarantee convergence on the trajectory. Then given the actual position $\vec{q}$, actual speed $\dot{\vec{q}}$ and desired acceleration $\ddot{\vec{q}}_r$ compute the optimal gear-ratios $R^*$, as described previously for a known trajectory, and apply the corresponding torques $\vec{\tau}^*$. The salient feature of the R* controller is that the optimal gear-ratio is selected based on state-feedback, i.e. even in situations not foreseen in the planner that generated the nominal trajectory. For instance, if a disturbance pushes the robot in a state where the robot faces a large gravitational force requiring a large gear-ratio, the controller will automatically select it. 



\subsubsection{ \color{review_color} Robust Control}
\label{sec:uncertainty}

{ \color{review_color}
In general Computed Torque Control is susceptible to modeling uncertainties and disturbances. This section presents two approaches to improving robustness: Adaptation and Sliding Mode Control. 

\paragraph{Adaptation}
If the uncertainty is structured as unknown model parameters in the extrinsic dynamics, the term represented by $\vec{\tau}_E$, then traditional adaptation schemes can be used for estimating the unknown parameters. Then, if adaptation converge to the correct computed torque, then the computed best gear-ratios will also converge to the true optimal gear-ratios:
\begin{align}
	\hat{\vec{\tau}}_E \rightarrow \vec{\tau}_E 
    \quad \Rightarrow \quad 
    \hat{R}^* \rightarrow R^*
 \label{eq:adapt}
\end{align}

\paragraph{Sliding Mode Control}

Alternatively, if the uncertainty (disturbances, noise and modeling errors) is bounded, Sliding Mode Control can be applied to improving robustness. 
Introducing intermediary variables defined as:
\begin{align}
	\vec{q}_e        = \vec{q} - \vec{q}_d  \quad \quad
	\vec{s}          = \dot{\vec{q}}_e + \lambda \vec{q}_e \quad \quad
  \vec{\ddot{q}}_r = \ddot{\vec{q}}_d - \dot{\vec{q}}_e
 \label{eq:slidingvar}
\end{align}

the computed torque, eq. \eqref{eq:eom_ndof2}, is modified to the following sliding mode control law: 
\begin{align}
	\vec{\tau} &=  R^{-1} 
	\vec{\tau}_{E}(\ddot{\vec{q}}_r,\dot{\vec{q}},\vec{q}) 
	+ R 
	\vec{\tau}_{I}(\ddot{\vec{q}}_r,\dot{\vec{q}})
    + R^{-1} G sgn( \vec{s} ) 
 \label{eq:slidingctl}
\end{align}
Convergence to the desired trajectory, despite the uncertainty, can be guaranteed using the Lyapunov function $V=\vec{s}^T \vec{s}$. The sliding condition is guarantee \cite{asada_robot_1986}, for any selected gear-ratios, if the discontinuous gain is set to:
\begin{align}
	G &= \left[ H + R^T I_a R \right] diag( \vec{k} ) \\ k_{i} &> \max \left| \left(  \left[ H + R^T I_a R \right]^{-1} \vec{d} \right)_{i} \right|
 \label{eq:slidingcond}
\end{align}
If the gear-ratio is selected to minimize the sliding mode torque, eq. \eqref{eq:slidingctl}, instead of the computed torque, eq. \eqref{eq:eom_ndof2}, then naturally larger gear-ratio  will be selected in response to large uncertainty. For instance, for a 1-DoF case:
\begin{align}
	R^{*} &= \operatornamewithlimits{argmin}\limits_{R} \left[ \tau^2 \right] = \sqrt{ \left | \frac{\tau_{E}(\ddot{q}_r,\dot{q},q) + | d |_{max} \, sgn( s ) }{\tau_{I}(\ddot{q}_r,\dot{q})} \right |   } 
\label{eq:rstar_sliding}
\end{align}
Hence, if no disturbance is expected ($| d |_{max}=0$) gear-ratio selection is unaffected, but when large disturbances are expected ( $| d |_{max}$ is large) torque minimization naturally leads to selecting large gear-ratios.
}

\subsubsection{Numerical Optimization}
If the gear-ratios are limited to a few discrete choices, the gear-ratio matrix $R$ is not diagonal or intrinsic actuator dynamics is non-linear, then the optimization must be computed numerically. The minimum needed is to have a model of the inverse dynamic, which could include any non-linearity, in the form $\vec{\tau}  = f( \vec{\ddot{q}} , \vec{\dot{q}} , \vec{q}  , R )$. In the situation of discrete gear-ratios, this lead to a combinatorial optimization problems, adding a constraint on the gear selection in the form $R \in \{R_1,R_2, ... , R_l\} $ However, if the number of options $l$ is reasonably small, then every possible options can be computed quickly. 

\subsubsection{Heuristic approach to minimize switching}
\label{sec:HeuristicApproachToMinimizeSwitching}
Because the control effort during gear-shifts is neglected in the model, using the proposed controller can lead to rapid switching between gear-ratios in certain situations. To avoid this undesirable behavior, it is proposed to add hysteresis to the controller: the gear-ratio is only changed if the difference of computed torque, between using the optimal and the previously selected gear-ratio, is greater than a minimum torque threshold, and also if the elapsed time since the last change is greater than a minimum delay. { \color{review_color}Ideally, the trade-off between fast and slow switching should be set based on a measure of how costly transitions are, which would be specific to the type of mechanism used by the VGA. }

\subsection{Examples of gear-selection in simple 1-DoF situations}
\label{sec:Examples}
\paragraph{Acceleration from rest} 
When a robot accelerates from rest with no viscous forces, in a configuration where no gravity acts, the problem is reduced to impedance matching for two inertial loads. The optimal gear-ratio minimizing the torque for a given acceleration is the one for which the load inertia and the motor reflected inertia are the same.
\begin{align}
	R^{*}  = \sqrt{ \left | \frac{H \ddot{q}_r }{ I \ddot{q}_r } \right |   } = \sqrt{ \frac{H}{I}}
 \label{eq:impmatching}
\end{align}
\paragraph{Supporting gravity without moving}
In the situation where the robot is not moving and fighting against gravity, the largest possible gear-ratio is the optimal choice. 
\begin{align}
	R^{*}  = \sqrt{ \left | \frac{ g }{ 0 } \right |   } \rightarrow \infty
 \label{eq:gravrejection}
\end{align}
{ \color{review_color}
\paragraph{Resisting disturbances}
In the situation where there is no gravity and the robot is not moving, but disturbances are expected, minimizing the sliding mode torque (eq. \eqref{eq:rstar_sliding}) would also lead to the conclusion that the largest possible gear-ratio is the optimal choice.}
\begin{align}
	R^{*}  = \sqrt{ \left | \frac{ 0 + | d |_{max} \, sgn( s ) }{ 0 } \right |   }  = \sqrt{  \frac{| d |_{max} }{ 0 } }\rightarrow \infty
 \label{eq:gravrejection}
\end{align}

\section{Simulation Results}
\label{sec:sim}

\begin{figure}[htp]
	\centering
		\includegraphics[width=0.40\textwidth]{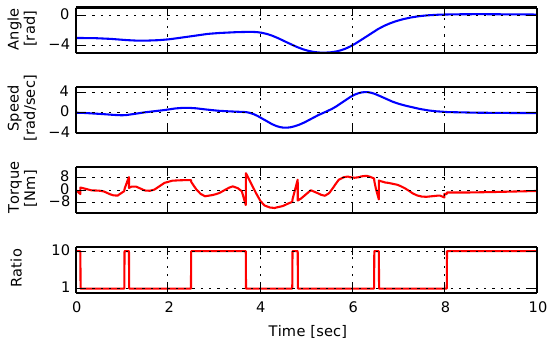}
	\caption{1-DoF robot simulation: states and inputs trajectory}
	\label{fig:sim1}
\end{figure}

In this section, the advantages of dynamically changing the gear-ratios, using the R* computed torque controller, are illustrated using simulations of two robots: first a 1-DoF pendulum, then a 3-DoF arm. Both robots are considered having VGA with two possible gear-ratios: 1:1 or 1:10. Reference low-torque trajectories to reach target positions are computed offline using a sample-based search algorithm \cite{lavalle_planning_2006}. 
The first simulated experiment uses the robot of Fig. \ref{fig:bigpicture}, but here considering dissipative forces in the actuators, with the task of reaching the up-right position starting at the bottom. Fig. \ref{fig:sim1} shows the robot tracking the reference low-torque trajectory, where at first the robot accumulates energy, using the 1:1 gear-ratio, and then finishes the motion using the 1:10 gear-ratio. When the gravitational forces are pushing advantageous toward the trajectory the controller select the 1:1 gear-ratio, but when it is advantageous to fight the intrinsic actuator dynamics instead, the 1:10 gear-ratio is selected. 

\begin{figure}[htp]
	\centering
		\includegraphics[width=0.40\textwidth]{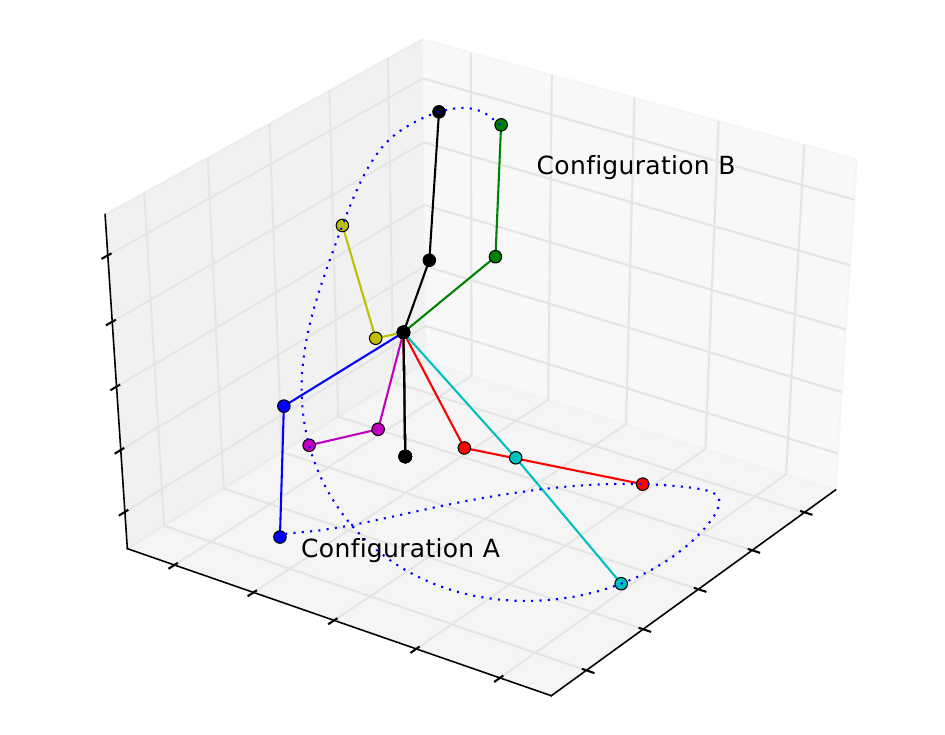}
	\caption{ 3-DoF robot simulation: 3D trajectory }
	\label{fig:3d_traj}
\end{figure}

\begin{figure}[htp]
	\centering
		\includegraphics[width=0.40\textwidth]{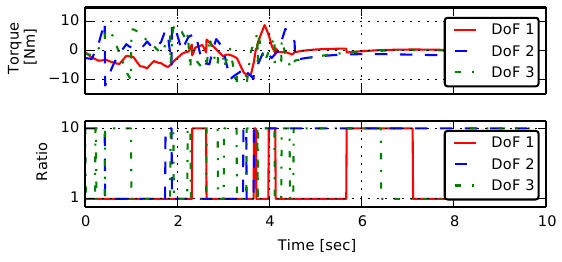}
	\caption{ 3-DoF robot simulation: control inputs trajectory}
	\label{fig:3d_u}
	\vspace{-10pt}
\end{figure}

%
%
%
%
%
In the second experiment, a 3-DoF manipulator is tasked with going from configuration A to configuration B with the 3D trajectory shown at Fig. \ref{fig:3d_traj}. For this robot the controller is actively selecting the best gear-ratios matrix $R$ out of the possible $2^3=8$ options. Fig. \ref{fig:3d_u} shows the control inputs activity. During the initial falling-down phase, at around $t=1$, the robot is using 1:1 gear-ratios for all actuators, leveraging gravitational torques. In contrast, during the final lifting phase, at around $t=6$, the robot is using 1:10 gear-ratios for all actuators. 
%
%
To evaluate the performance gain of actively changing the gear-ratio, simulations with fixed gear-ratios are conducted where a regular computed torque controller tracks the same trajectories. Results are summarized in TABLE \ref{tab:MaximumTorqueComparison}, in terms of maximum absolute torque, which relates to the required size and weight of motors, and integral of torque squared, which relates to power consumption. Active gear-ratio selection is found to greatly improve both metrics, especially for the 3-DoF robot trajectory where the arm must both achieve high-speeds and also sustain a constant gravitational load at the final configuration. Note that in those simulations high-velocity with 1:10 reductions is inhibited by friction in the motors, no maximum rotor velocity is enforced. For the 3-DoF trajectory, active gear-shifting is found to reduce the maximum torque required by a factor two and the integral of the torque square by a factor 10, compared to any of the fixed-gear options. Those results show the potential of using variable gear-ratio transmissions for huge improvements in terms of actuator size and power consumption. Moreover, here in the simulations, the load was always the same manipulator in different dynamic situations. As discussed in section \ref{sec:app}, if the load dynamics is radically changing because of different contact situations with the environment, the performance gain of changing the gear-ratio could be even greater. 
\begin{table}[htp]
	\centering
		\begin{tabular}{ c c c c }
		\hline
		     & Fixed gear & Fixed gear & Active gearshifting \\
			& 1:1 &  1:10 &  1:1 or 1:10 \\
		\hline \hline
		\multicolumn{4}{c}{ Max Absolute Torque [Nm] } \\
		\hline \hline
		1-link robot  & 15 & 88 & 12 \\	
		3-link robot  & 24 & 42 & 12 \\	
		\hline \hline
		\multicolumn{4}{c}{ Torque squared integral $\int{ ( \vec{\tau}^T \vec{\tau} ) dt }$ } \\
		\hline \hline
		1-link robot  & 377  & 8133 & 226  \\	
		3-link robot  & 2774 & 3617 & 295  \\	
		\hline \hline
		\end{tabular}
	\caption{Required torque comparison}
	\label{tab:MaximumTorqueComparison}
	\vspace{-20pt}
\end{table}
%
%

\section{Experiment results}
\label{sec:ev}

This section presents a novel robotic arm with variable gear-ratio actuators and an experimental validation of the R* computed torque controller and the sliding mode version.

\subsection{Prototype presentation}
\label{sec:proto}

\begin{figure}[htp]
	\centering
		\includegraphics[width=0.40\textwidth]{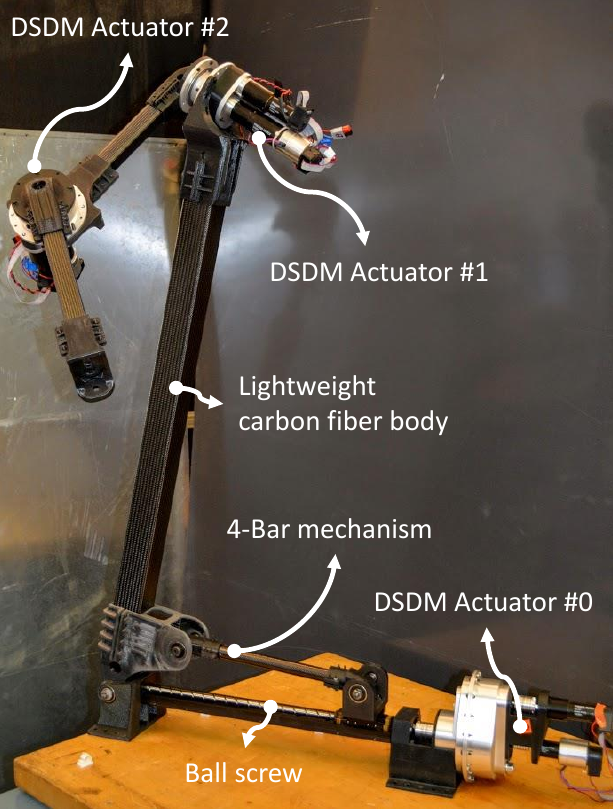}
	\caption{Prototype 3-DoF arm with variable gear-ratio actuators}
	\label{fig:arm_proto}
\end{figure}

\begin{figure}[htp]
	\centering
		\includegraphics[width=0.40\textwidth]{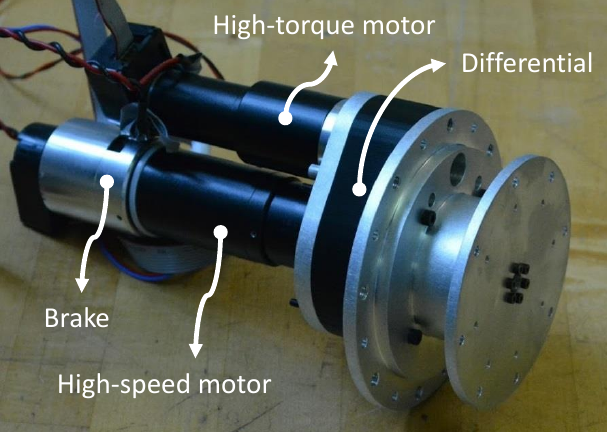}
	\caption{Two speed (1:23 or 1:474) variable gear-ratio actuator prototype}
	\label{fig:dsdm_cad}
\end{figure}

Shown in Fig. \ref{fig:arm_proto}, a 3-DoF robotic arm using variable gear-ratio actuators has been designed and custom built. The variable gear-ratio actuators consist of dual-speed dual-motor (DSDM) actuators, as shown in Fig. \ref{fig:dsdm_cad}. Those actuators have two discrete operating modes, one for high-speed operation, and the other for low-speed, high-load operation, where the net gear-ratios are about 20-times different \cite{girard_two-speed_2015}. Also, the dual-motor architecture has the advantages of allowing fast and seamless gear-shifts, supporting the modeling assumption discussed in sec.\ref{sec:mod}. 
%

\subsection{Controller implementation}
\label{sec:ctl_imp}

The control algorithms are implemented on a computer using ROS \cite{quigley_ros:_2009}, the trajectory generation algorithm and the R* Computed Torque controller are written in \textit{Python}. The computer is communicating over USB with \textit{Flexsea} motor-drivers \cite{duval_flexsea-execute:_2016} that handle the low-level current loops. The trajectory is generated offline in advance and loaded in memory upon initialization. The main R* control loop is running at a 500 Hz sampling rate. The low-level actuator controllers, receiving the torque and gear-ratio set-points, communicating with both motor drivers and handling the gear-shifting process (see Fig. \ref{fig:control_achitecture}) use the algorithm described in \cite{girard_two-speed_2015}. Transition from one gear-ratio to another are found to be consistently under 50 ms.

\subsection{Experiments}
\label{sec:exp}

\begin{figure}[htp]
	\centering
		\includegraphics[width=0.40\textwidth]{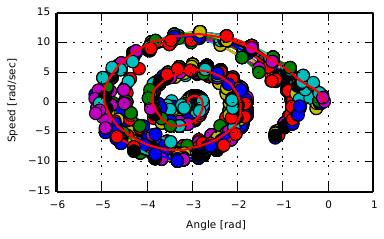}
		\vspace{-10pt}
	\caption{Low torque reference trajectory found using a search algorithm}
	\label{fig:exp_rrt}
\end{figure}

\begin{figure}[htp]
	\centering
		\includegraphics[width=0.45\textwidth]{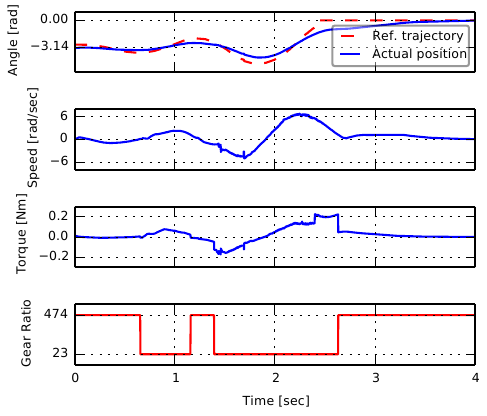}
	\caption{Experiment with the R* Computed Torque Controller}
	\label{fig:exp_traj}
\end{figure}

First a trajectory following experiments using the last DoF of the robot only is presented. A 1.5 Kg load is mounted on the end-effector, and the task is to bring it from the bottom position ($q=-\pi$) to the up-right position ($q=0$) using as little torques as possible. A search algorithm is used to find a low torque trajectory reaching the goal, see Fig. \ref{fig:exp_rrt}. Then the R* Computed Torque Controller is used to track the reference trajectory. The experimental results are shown in Fig. \ref{fig:exp_traj}.
Results show that the robot is using its 1:23 gear-ratio to accumulate kinetic energy by swinging the arm link back and forth. Also the R* controller selects the 1:474 gear-ratio automatically to attenuate the load dynamics, when the actuator has to force the robot to stay with the trajectory.  Interestingly, the reference trajectory was planned so the robot would accumulate enough kinetic energy to swing straight up with the last swing. However, in the experiment, the dissipative forces are greater than anticipated by the planner, and the last swing is too small (the robot almost stop at $q=-0.9$ at $t=2.6$ in Fig. \ref{fig:exp_traj}). Then, the R* controller automatically engage the large 1:474 gear-ratio, to continue converging on the desired trajectory with much smaller torques than those required if keeping using the 1:23 gear-ratio in this situation (no momentum and a large gravitational force to overpower). This illustrates that including the gear-ratio selection in the feedback loop also increase the robustness of the system. Without the 1:474 gear-ratio option, tracking would have failed as the computed torque with 1:23 in this situation was greater than the maximum allowable motor torque.

{ \color{review_color}
Fig. \ref{fig:rob} shows four additional experiments for demonstrating how disturbance rejection can be improved by using the sliding mode version of the R* controller. Here the controller is only given a simple fixed point-target in all cases. First, when a low uncertainty bound is given to the controller, the robot can reach its target when unloaded (a) but failed when an unknown (to the controller) 0.4 Kg load is added to the end-effector (b). However, when a larger uncertainty bound is given, the robot can reach its target in both cases, unloaded (c) and loaded (d). Note that the discontinuous torque required to guarantee convergence despite disturbances is greatly reduced when down-shifting to a large gear-ratio at low speeds. For an improved performance, smoothing techniques should be implemented to avoid exciting the unmodeled high-frequency modes. Videos of all the experiments discussed above, and additional demonstrations are available in the media attachment.

\begin{figure}[htp]
        \centering
				\hspace{-10pt}
				\subfloat[]{ 
				\includegraphics[width=0.15\textwidth]{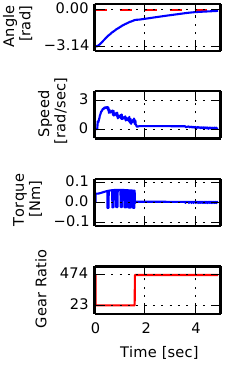} }
				\hspace{-5pt}
        \subfloat[]{ 
				\includegraphics[width=0.10\textwidth]{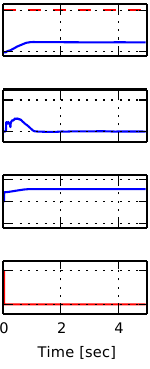} }
				\hspace{-5pt}
				\subfloat[]{ 
				\includegraphics[width=0.10\textwidth]{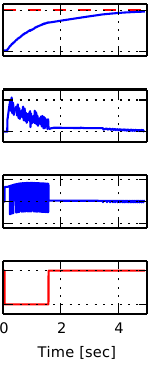} }
				\hspace{-5pt}
				\subfloat[]{ 
				\includegraphics[width=0.10\textwidth]{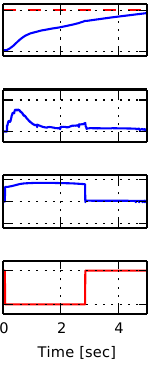} }
        \caption{Experiments with sliding mode version of the R* controller }
				\label{fig:rob}
\end{figure}
}

\section{CONCLUSION AND OUTREACH}

This paper explored the idea of dynamically changing the gear-ratio of an actuator to minimize necessary motor torques, minimize the power consumption, and modulate the output impedance of a robot, which would benefit greatly many mobile robotics applications. This paper main contribution is a model-based control algorithm to automatically select the best gear-ratios for any dynamic situation generalized to n-DoF fully actuated robotic systems. Also, a custom lightweight 3-DoF robotic arm using variable gear-ratio actuators with two discrete gear-ratio options is presented. The advantages of gear-shifts were illustrated using simulations and experiments. 

Since the idea of dynamically changing the gear-ratio of a robot actuators can lead to potentially significant performance gains and is relativity unexplored compared to other topics in the field of robotics, there is many aspects that should be investigated further: how to best optimize trajectories for this kind of hybrid system, how to best use the ability to change the reflected mass of the robot for interaction tasks, how to use gear-shifting to improve legged locomotion, {\color{review_color}addressing applications with fast dynamics where gear-shift transition delays cannot be neglected, etc.}



\bibliographystyle{IEEEtran}
\bibliography{main}

\end{document}